\documentclass[10pt,twocolumn,letterpaper]{article}

\usepackage[pagenumbers]{cvpr}

\usepackage{graphicx}
\usepackage{amsmath}
\usepackage{amssymb}
\usepackage{booktabs}

\usepackage{epigraph}

\usepackage{url}            %
\usepackage{amsfonts}       %
\usepackage{nicefrac}       %
\usepackage{xcolor}         %

\usepackage{xspace}
\usepackage{multirow}
\usepackage{mathtools}
\graphicspath{ {./images/} }

\usepackage[pagebackref,breaklinks,colorlinks]{hyperref}

\usepackage[capitalize]{cleveref}
\crefname{section}{Sec.}{Secs.}
\Crefname{section}{Section}{Sections}
\Crefname{table}{Table}{Tables}
\crefname{table}{Tab.}{Tabs.}

\usepackage{soul}

\definecolor{ubpubColor}{rgb}{0.43, 0.5, 0.5}
\definecolor{backboneColor}{rgb}{0.423, 0.325, 0.365}
\definecolor{fpnColor}{rgb}{0.255, 0.498, 0.416}

\newcommand{\PAR}[1]{\vskip4pt \noindent {\bf #1~}}

\newtoggle{comments}
\toggletrue{comments}
\iftoggle{comments}{%
\newcommand{\lau}[1]{\textcolor{magenta}{\textbf{Laura: }{#1}}}
\newcommand{\man}[1]{\textcolor{brown}{\textbf{Manuel: }{#1}}}
\newcommand{\alj}[1]{{\leavevmode\color{blue}#1}}
\newcommand{\achal}[1]{{\leavevmode\color{orange}[\textbf{Achal:} #1]}}
\newcommand{\cs}[1]{\textcolor{OliveGreen}{\textbf{CS: }{#1}}}

\newcommand{\newtext}[1]{{\leavevmode#1}}

}{%
\newcommand{\lau}[1]{}
\newcommand{\man}[1]{}
\newcommand{\alj}[1]{}
\newcommand{\achal}[1]{}
\newcommand{\cs}[1]{}
\newcommand{\deva}[1]{}
\newcommand{\bigb}[1]{}
\newcommand{\bastian}[1]{}
\newcommand{\laura}[1]{}
\newcommand{\jono}[1]{}
}
\newcommand{\embd}[1]{$\mathlarger{\mathlarger{\varepsilon}}$}

\def\cvprPaperID{2220} %
\def\confName{CVPR}
\def\confYear{2022}

\begin{document}

\title{Text2Pos: Text-to-Point-Cloud Cross-Modal Localization}

\author{Manuel Kolmet
\quad
Qunjie Zhou
\quad
Aljoša Ošep
\quad
Laura Leal-Taix\'{e}\\
Technical University of Munich, Germany\\
{\tt\small \{manuel.kolmet, qunjie.zhou, aljosa.osep, leal.taixe\}@tum.de} \\
\small{\texttt{
\href{https://text2pos.github.io}{text2pos.github.io}
}}
}

\twocolumn[{%
\renewcommand\twocolumn[1][]{#1}%
\maketitle
\begin{center}
\vspace{-0.5cm}
\includegraphics[width=0.999\linewidth]{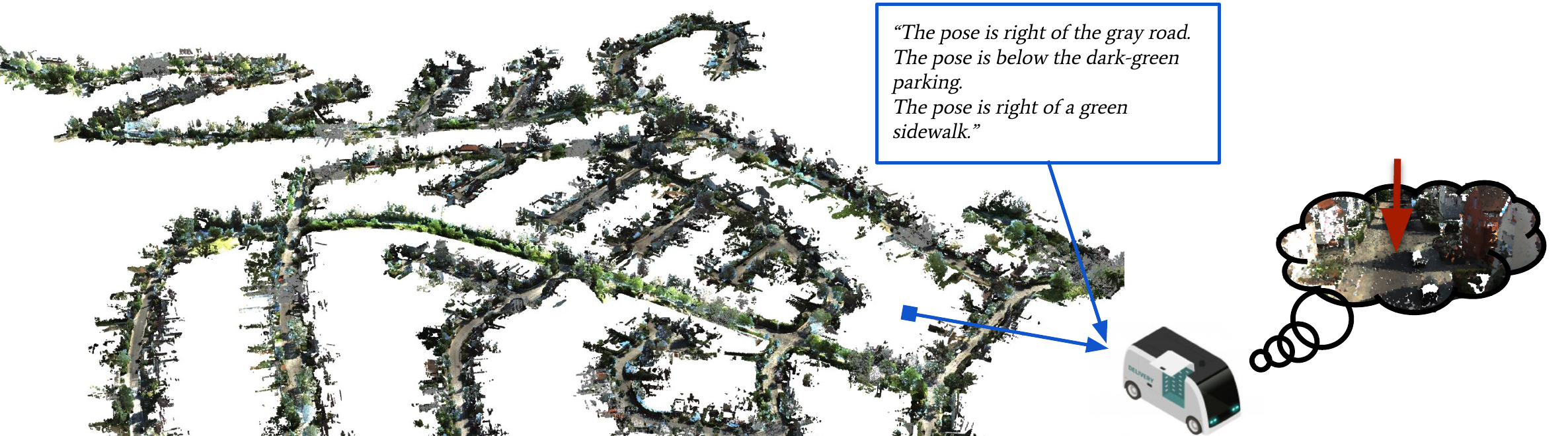}
\vspace{-14pt}
\captionof{figure}{We propose Text2Pos for city-scale position localization based on textual descriptions. Given a point cloud that represents our surroundings and a query position description, Text2Pos provides the most-likely estimate of the described position within that map.}
\label{fig:teaser}
\end{center}%
}]
\begin{abstract}
\vspace{-0.3cm}
    Natural language-based communication with mobile devices and home appliances is becoming increasingly popular and has the potential to become natural for communicating with mobile robots in the future. 
    Towards this goal, we investigate cross-modal text-to-point-cloud localization that will allow us to specify, for example, a vehicle pick-up or goods delivery location.  
    In particular, we propose Text2Pos, a cross-modal localization module that learns to align textual descriptions with localization cues in a coarse-to-fine manner.
    Given a point cloud of the environment, Text2Pos locates a position that is specified via a natural language-based description of the immediate surroundings. To train Text2Pos and study its performance, we construct KITTI360Pose, the first dataset for this task based on the recently introduced KITTI360 dataset. 
    Our experiments show that we can localize $65\%$ of textual queries within $15m$ distance to query locations for top-10 retrieved locations. This is a starting point that we hope will spark future developments towards language-based navigation. 
\end{abstract}
\vspace{-0.6cm}

\epigraph{“Alexa, hand me over my special delivery at the sidewalk in front of the yellow building next to the blue bus stop.” }{\textit{Authors of this paper, the future. Hopefully.}}

\section{Introduction}
\label{sec:intro}

\newtext{Future mobile robots, such as autonomous vehicles and delivery drones, will need to cooperate with humans to coordinate actions and plan their trajectories. In this paper we tackle large scale position localization of the target position based on natural-language-based position descriptions, as needed for, \eg, for goods delivery or for vehicle pickup.}

For self-localization within a map, mobile agents rely on visual localization methods~\cite{Arandjelovic2016CVPR, Kendall2015ICCV, Zhou2020ICRA, Sattler2016TPAMI, Shotton2013CVPR}. These methods match observed images either to a database of geo-tagged images~\cite{Arandjelovic2016CVPR, Torii2015CVPR, hausler2021patch} or point-cloud-based maps~\cite{Sarlin2019CVPR, Sattler2016TPAMI, Schonberger2018CVPR}, often obtained using structure-from-motion techniques~\cite{Schonberger2016CVPR, schonberger2016ECCV}. 
\newtext{By contrast, in this paper, we study \textit{language-based localization of any location}, which, importantly, \textit{does not require the user to be physically present at the target location.} This would, for example, allow us to explain the pick-up position or delivery location through text/voice to a robo-taxi via natural language based communication, that is preferable to humans.} 
\newtext{Our method can also be seen as} complementary to GPS localization methods, \eg, when a GPS tag is too coarse, unavailable, or language-based communication is more convenient.

As the main contribution of this paper, we formalize the task of language-based localization and provide the first dataset and methods for this task. In this problem setting, we assume an intelligent agent is given access to the map of the environment that comes in the form of a 3D point cloud \newtext{and object instance labels}. 
While there are several ways of acquiring point clouds, we rely on LiDAR point clouds, readily available in modern automotive~\cite{Caesar19arXiv,sun20CVPR,Xie2016CVPR} and urban~\cite{martin21PAMI} datasets, 
On the query side, we assume a textual description of the query position surroundings, such as the one shown in Fig.~\ref{fig:teaser}. The task is then to provide the most likely position estimate based on this query. 

To study this challenging problem, we need a dataset that (i) provides a point-cloud-based representation of the environment and (ii) provides labels in the form of query positions and their corresponding textual descriptions, extracted from their immediate surroundings. We build on the recently proposed KITTI360 dataset~\cite{Xie2016CVPR}, which provides nine scenes (city districts), covering 80 km of driving data. Importantly, this dataset provides semantic and instance-level annotations of the point cloud, which we use to automate the generation of query position descriptions. We obtain the \textit{KITTI360Pose} dataset by randomly sampling query positions and by generating corresponding textual descriptions. 
For each query position we automatically generate multiple descriptions based on a natural language template that specifies spatial relations of surrounding instances to the query position, together with their semantic classes and appearance. We generate $43,381$ such descriptions for $14,934$ sampled positions, which we split by scene (representing a city district) to obtain our train/test splits. 

We use this dataset to train and evaluate our proposed Text2Pos model that performs coarse-to-fine localization. 
In the coarse localization step, we retrieve sub-regions of the map that likely contain our target position.
To this end, our network learns to align the encoded query with the point clouds, representing these sub-regions. 
We finally refine the position estimate within retrieved candidate regions using our matching-based fine localization module.
Our experiments show that we can localize such randomly generated positions within KITTI360 scenes with 65\% \newtext{recall} for top-10 queries, demonstrating that localizing positions based on textual descriptions is feasible.  

In summary, our \textbf{main contributions} are:
we (i) introduce and formalize the task of 3D point-cloud based localization based on textual descriptions. To this end, we (ii) provide \textit{KITTI360Pose}, the first public dataset for this task, based on the KITTI360 dataset, together with our method for automated mining of positions and corresponding textual descriptions.
We (iii) provide a coarse-to-fine baseline model for the localization task, that learns to align objects which are mentioned in the text with object instances in the point cloud and thoroughly evaluate and ablate the performance on this challenging new task. 
We believe this work is the first step towards natural language-based communication with future mobile robots, such as delivery drones and self-driving taxis. 

\section{Related work}

\PAR{Vision-based localization.} Related to our problem is the task of visual localization \cite{Brachmann2017CVPR, Arandjelovic2016CVPR, Torii2015CVPR, Kendall2015ICCV, Zhou2020ICRA, Sarlin2019CVPR, Sattler2016TPAMI, Shotton2013CVPR, Valentin2015CVPR, Radwan2018RAL, hausler2021patch}, which means estimating a precise pose based on an observed image or image sequence. 
Existing methods commonly adopt a two-stage coarse-to-fine localization pipeline~\cite{Sattler2016TPAMI, Zhou2020ICRA, Sarlin2019CVPR}.
Given the query image, a \textit{coarse} step firstly finds a subset of images with aligning views using image retrieval techniques~\cite{Arandjelovic2016CVPR, hausler2021patch, Torii2015CVPR}.
Then, the \textit{fine} step establishes 2D-2D correspondences between pixels of the query and the retrieved images based on the visual descriptors. These can further be used to obtain 2D-3D correspondences between the query and a 3D map, usually obtained using structure-from-motion techniques.
Finally, the camera poses can be computed using either a set of 2D-2D correspondences~\cite{Zhou2020ICRA} or 2D-3D correspondences ~\cite{Sarlin2019CVPR, Sattler2016TPAMI}.
Our method follows the coarse-to-fine localization scheme by first localizing a coarse cell, containing the objects described by the query text, followed by a more precise pose estimate within the coarse cell. 
Compared to matching between visual features, our method needs to implicitly learn to align two different modalities: text and 3D point clouds. \newtext{By contrast to visual localization, commonly used for robot \textit{self-localization}, we tackle linguistic localization, intended to specify an arbitrary \textit{target location}}.

\PAR{2D vision and language.} Vision and language understanding has been widely investigated in tasks such as image captioning~\cite{Karpathy2015CVPR, Vinyals2015CVPR, Xu2015ICML, Lu2017CVPR}, visual question answering (VQA)\cite{Antol2015ICCV, Wang2017TPAMI} and visual grounding \cite{Yu2018CVPR, Mao2016CVPR, Hu2016ECCV, Kazemzadeh2014EMNLP}, the task of localizing visual elements in the images that are linguistically described by the query text.
Visual and linguistic perceptions are combined to assist the task of robot navigation from room to room under building-scale environments~\cite{Anderson2018CVPR}. 
The ALFRED benchmark~\cite{Shridhar2020CVPR} was later released to encourage research on connecting language to a series of human daily tasks in an interactive visual 3D environment.
Closer to our task is text-to-image retrieval~\cite{Wang2016CVPR, Huang2018CVPR, Messina2021ICPR, Wang2020ECCV, Messina2020TOMM}, where text descriptors are learned to match corresponding image descriptors.
This usually requires the model to reason about the relationship between a set of words and image regions and match a word to its corresponding image region \cite{Messina2021ICPR}.
The main difference between our approach and previous work is that we match from text to point clouds instead of images.
In the coarse localization stage, our method first matches a sequence of texts to a cell which represents a region in the scene and contains a set of objects that are then matched to individual textual object hints. 

\PAR{3D vision and language.} Motivated by the 3D world we live in, recent work explores the potential of 3D vision and language understanding on the tasks of 3D shape generation~\cite{Chen2018ACCV} and language grounding of 3D objects ~\cite{Prabhudesai2019arXiv, Chen2020ECCV, Achlioptas2020ECCV, Yuan2021Arvix, Feng2021Arxiv}. The method by~\cite{Prabhudesai2019arXiv} embodies language grounding implicitly on 3D visual features and predicts 3D bounding boxes for target objects of primitive shapes of different colors. 
ScanRefer~\cite{Chen2020ECCV} localizes 3D objects referred by the query descriptions in real-life indoor scenes.
ReferIt3D~\cite{Achlioptas2020ECCV} tackles a similar task, but assumes to be given segmented object instances in a room and focuses on identifying the referred object among instances of the same fine-grained category.
InstanceRefer~\cite{Yuan2021Arvix} improves their performance by using a 3D panoptic segmentation backbone, guiding the model to capture multi-level visual context. 
Recent work by~\cite{Feng2021Arxiv} proposes several graph modules that aid learning of the contextual information for both visual and language domains. 

Similar to our work, these methods localize regions in 3D point clouds based on textual queries. 
However, this is different to our proposed city-scale text-based position localization task in several aspects:
For 3D object reference, a model needs to learn to align a natural-language descriptor to one of the objects in the scene. 
Different to that, our model needs to learn to interpret a composition of objects as a location and distinguish it from other possible locations, since there is no explicit visual notion of a position. 
Additional challenges stem from the fact that we are targeting large (city) scale outdoor scene localization. This is challenging due to memory constraints of modern GPUs, which can fit only a very small portion of a city scale point cloud. Furthermore, outdoor regions are less diverse in terms of semantics compared to cluttered man-made indoor environments~\cite{minaee2021image}, making it difficult to leverage semantic instances to obtain a unique position signature. 
In summary, our work serves as the first attempt to tackle this challenging task and opens the door to natural language based localization for the 3D vision and language community.

\section{The \textit{KITTI360Pose} dataset}
\label{sec:dataset}

To tackle language-based position localization in large-scale environments such as urban cities, we (i) need a large-scale dataset that provides point clouds representing real-world cities and (ii) a large set of position-text pairs to train and evaluate our models. 
To this date, no such dataset exists, and hand-annotating textual queries would be prohibitively expensive. 

The recently introduced KITTI360~\cite{Xie2016CVPR}\footnote{Available under Creative Commons  \href{https://creativecommons.org/licenses/by-nc-sa/3.0/}{Attribution-NonCommercial-ShareAlike 3.0 license}. This dataset contains no personally identifiable information or offensive content.} dataset provides nine static scenes that represent different districts of the city of Karlsruhe, covering in total over $80 km$ of driving distance. These scenes were obtained by registering LiDAR scene scans using LiDAR SLAM methods (\eg,~\cite{behley18rss}). These point clouds would be suitable for studying this problem; however, the dataset does not provide textual descriptions of positions. Luckily, KITTI360 provides object instance labels for static (\eg, building, traffic light, garage) and dynamic (\eg, person, car, bicycle) object instances and semantic labels for the \textit{stuff} classes (\eg, road, vegetation, wall). 
In the following, we utilize these object instance and semantic labels to automatize the generation of position-description query pairs. We use these to train our models and to benchmark large-scale cross-modal localization without manual annotation work.

In this study, we focus on point clouds, recorded by LiDAR sensors, readily available in modern automotive~\cite{Caesar19arXiv,sun20CVPR,Xie2016CVPR} and robotics~\cite{martin21PAMI} datasets. 
Our approach would also be applicable to point clouds obtained using structure-from-motion methods~\cite{Schonberger2016CVPR, schonberger2016ECCV} available in existing visual localization datasets~\cite{Kendall2015ICCV, Sattler2018CVPR}. However, such datasets currently do not contain appropriate instance annotations that we could utilize to automatize the generation of query positions. Finally, we note that indoor RGB-D datasets, such as~\cite{Dai2017CVPR, Chang2017matterport3d} do contain such object instance labels. However, in this paper, we explicitly aim to study large-scale localization, hence, focus on outdoor scenarios.

\begin{figure*}[t!]
  \centering
  \includegraphics[width=0.98\textwidth]{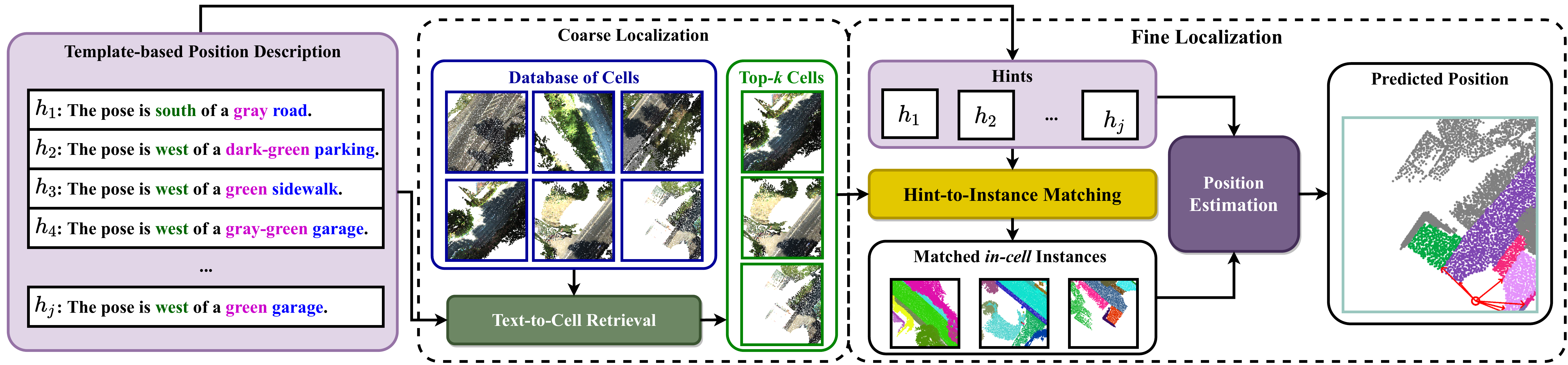}%
    \vspace{-7pt}
  \caption{\textbf{Text2Pos}. \textit{Coarse localization.} Given a template-based query position description, we first identify a set of coarse candidate locations (\ie, cells) that potentially contain the target position, which serves as the coarse localization of the query. This is achieved by retrieving \newtext{\textit{top-k}} nearest cells from our constructed database of cells using our text-to-cell retrieval model. \textit{Fine localization.} we then refine the pose within retrieved cells via our position refinement module.}
  \label{fig:pipeline}
\end{figure*}

\subsection{Dataset Generation}\label{sec:dataset_plc}

\PAR{Object instances.} Contrary to the majority of existing automotive datasets~\cite{sun20CVPR, Caesar19arXiv}, that focus on instance segmentation of dynamic objects such as \textit{cars} and \textit{pedestrians}, KITTI360 additionally provides object instance labels for several static classes, such as \textit{buildings} and \textit{traffic lights}, that provide a reliable cue for localization. In this work, we leverage static object instances to generate position queries and as cues for the position localization. 

In addition to labeled instances, we also further split certain \textit{stuff} classes and use the obtained clusters to generate descriptions. For example, the class \textit{vegetation} aggregates a large set of separate trees and bushes which could be specified as localization cues into a single object that spreads across the entire scene. In order to instead recover a set of separate and therefore localizable instances, we cluster all \textit{stuff} classes, such as \textit{vegetation}, \textit{fence} and \textit{wall} using the DBSCAN~\cite{Ester96KDD} algorithm.
We provide further details on the clustering procedure in the supplementary material. 

\PAR{Query generation.}
The next step is position-query pair generation. The aim here is to obtain a set of positions and corresponding texts that describe each position qualitatively based on the surrounding objects and their spatial relations in an automated fashion.
We start by sampling equidistant locations along recorded vehicle trajectories, readily available with maps. In the vicinity of each sampled location, we sample a fixed number of random locations (in practice, 4 or 8) to increase the number of positions.
We describe adjacent objects to each position based on their relative position, color and semantic class by generating textual descriptions based on a simple sentence template, describing the position in relation to adjacent objects. We extract the relative position and object color tag in an automated fashion directly from the point cloud. We detail the sentence generation in the supplementary.

In the following, we refer to one such generated sentence as a \textit{hint}.
A position description $\mathcal{T}$ is defined by a set of hints $\{h_i\}$ describing a given position, \ie, $\mathcal{T} \coloneqq \{h_i\}^{N_h}_{i=1}$ where $N_h$ is the number of hints per position description. 
The set of objects to describe a position with is obtained by selecting a set of \textit{k} objects close to the sampled query position. 
We retain only positions with at least $N_h$ objects in their vicinity. 
The reason for such well-structured queries is two-fold: (i) this allows us to investigate the problem without costly human annotations, and (ii) enables us to rigorously study cross-modal localization in a well-controlled setting based on explicit hint-to-object matching. We are confident that more complex language queries can be encoded and understood by building on recent developments in the field of natural language processing~\cite{brown2020language, devlin2018bert}. 

\PAR{KITTI360Pose dataset.}
Using the procedure described above, we generate the \textit{KITTI360Pose} dataset. 
In particular, we sample $14,934$ positions and generate up to three descriptions for each, totaling in $43,381$ position-query pairs. We use five scenes (districts) for training (covering in total $11.59 km^2$), one for model validation, and three for testing (covering in total $2.14 km^2$). An average district covers an area of $1.78 km^2$. 
In contrast, the Cambridge dataset~\cite{Kendall2015ICCV} covers an area of $0.063 km^2$, Oxford RobotCar and CMU Seasons~\cite{Sattler2018CVPR} cover $10 km$ and $8.5 km$ of driving distance (respectively), and Tokyo 24/7~\cite{Torii2015CVPR} covers the area of $2.56 km^2$. 
Descriptions of these positions are generated based on objects that fall within a $15 m$ radius for a sampled position. We provide additional details in the supplementary.

\section{Tex2Pos: A Baseline for Language-based Localization} 
\label{sec:method}
Given a textual position query, our goal is to localize an agent within a given point-cloud-based (Sec.~\ref{sec:dataset_plc}) map via 2D planar coordinates of its position \wrt the scene coordinate system. 
To this end, we propose the first text-based coarse-to-fine localization method, that we outline in Fig.~\ref{fig:pipeline}. 
Due to the large-scale nature of the problem, we follow a coarse-to-fine framework, well studied and proven successful in the field of large-scale visual localization~\cite{Sarlin2019CVPR, Zhou2020ICRA}. 
We first perform a coarse localization of the query where we discretize the search region into rectangular \newtext{cells} and retrieve the \newtext{\textit{top-k}} cells matched to the description from the database (Fig.~\ref{fig:pipeline}, \textit{left}). 
To refine this estimate, we match the visible 3D instances within a retrieved cell, to their corresponding referring hints in the textual description (Fig.~\ref{fig:pipeline}, \textit{right}). 
Finally, we obtain the position estimate from the set of instances identified by the text (Sec.~\ref{sec:method_fine}). 
We note that the coarse retrieval could in the future be replaced by hints such as street name or coarse address. This would require establishing an additional alignment between the 3D point cloud and city map, and remains our future work. 

\subsection{Coarse Localization}\label{sec:method_coarse}

Image retrieval techniques~\cite{Arandjelovic2016CVPR, hausler2021patch, Torii2015CVPR} are commonly used within visual localization pipelines to efficiently narrow down the search space~\cite{Sarlin2019CVPR,Taira2018CVPR, Zhou2020ICRA} or even provide a direct coarse position estimate~\cite{Sattler2019CVPR} of the query image~\cite{Sattler2018CVPR}.
Given a query image, its learned global descriptor is matched against the global descriptors extracted from a database of reference images to obtain its \newtext{\textit{top-k}} nearest reference images based on their descriptor distances.
We follow this general approach in our \textit{text-to-cell} cross-modal retrieval method. With this step, we aim to efficiently locate candidate regions within the map that could potentially contain our target position.

\PAR{Database construction.} 
As a pre-processing step, we divide point clouds, representing city districts, into rectangular \newtext{cells}.
We sample cells by sliding a $\mathcal{W} \times \mathcal{W}$ window with a stride of $\mathcal{S}$ horizontally and vertically over the scene, where the cell size $\mathcal{W}$ should be large enough to contain a certain number of instances, that can be used to describe a position.
The stride size $\mathcal{S}$ is picked to be smaller than the cell size to cover the whole scene area and to allow partially overlapping cells. 
We consider an instance $p_i$ to be inside a cell $\mathcal{C}$ if at least a third of its points lie within the cell, or if a minimal number of points (250 in practice) overlap with the cell -- this criterion is important for \textit{stuff} classes, such as \textit{tree} or \textit{building}. 
We name those instances \textit{in-cell} instances of that cell, \ie, $\mathcal{C}\coloneqq\{p_i\}^{N_p}_{i=1}$ where $N_p$ is the number of \textit{in-cell} instances per cell and varies for different cells.

\PAR{Text-to-cell retrieval.}
Given a position description, the task of our retrieval model is to identify its \newtext{\textit{top-k}} candidate cells that are likely to contain the described position location.
Compared to image-to-image retrieval, the model needs to learn to extract descriptors for inputs from two different modalities, \ie, text and point clouds, such that the two can be directly compared using Euclidean distance in the embedding space. 

\begin{figure}[t!]
  \centering
  \includegraphics[width=0.98\linewidth]{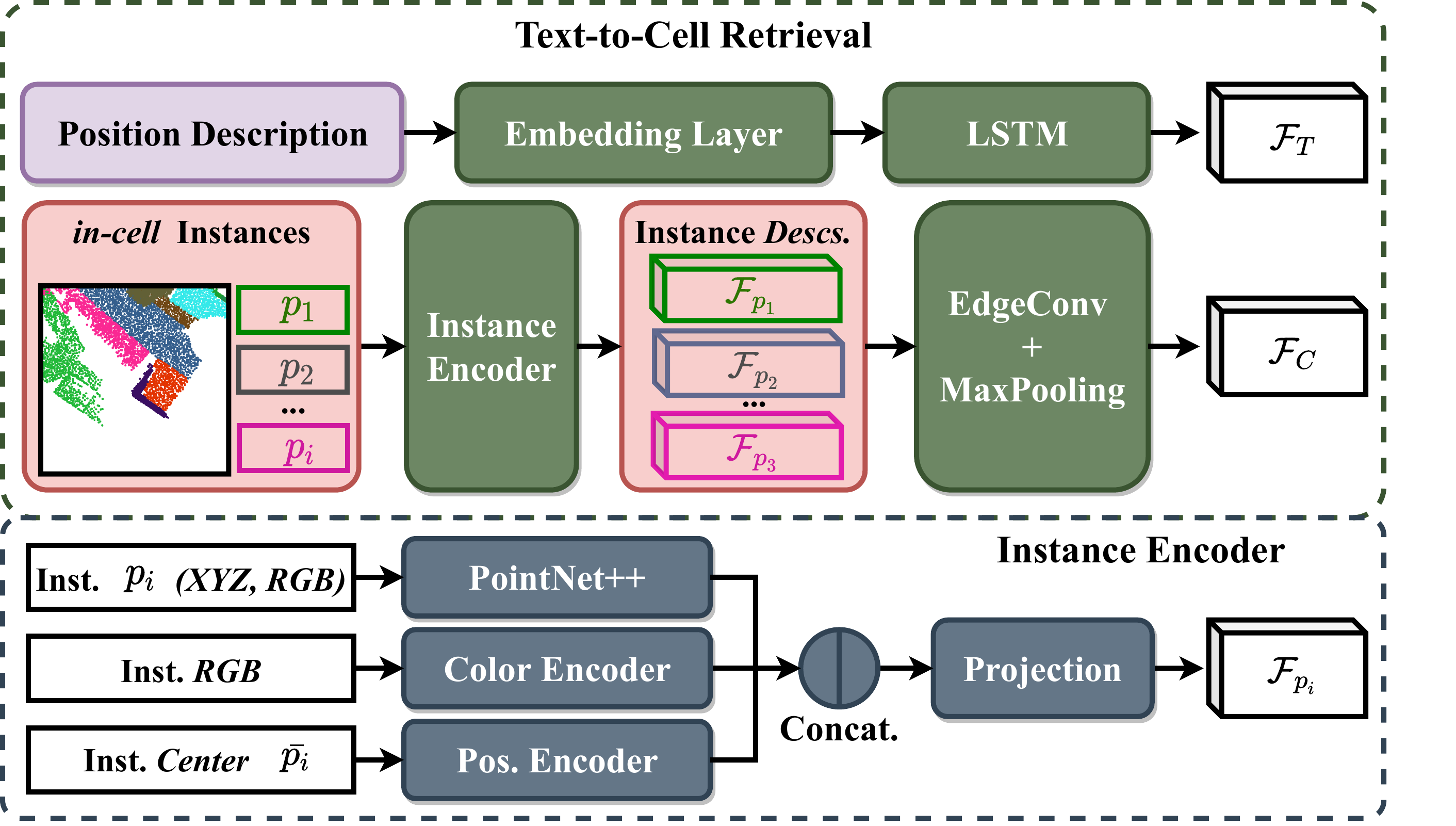}
    \vspace{-7pt}
  \caption{\textit{(top)} Query text encoder and cell encoder architecture, \textit{(bottom)} instance encoder architecture.}
  \label{fig:encoders}
\end{figure}

As shown in Fig.~\ref{fig:encoders} (\textit{top}), our retrieval network has two encoding branches to process a query position description $\mathcal{T}$ and a candidate cell $\mathcal{C}$.
The complete position description $\mathcal{T}$ is encoded into a global text descriptor $\mathcal{F}_T$ using a bidirectional LSTM cell~\cite{Hochreiter1997NC}. 
On the cell encoding side, we first extract a descriptor $\mathcal{F}_{p_i}$ for each \textit{in-cell} instance $p_i \in \mathcal{C}$. %
We aggregate \textit{in-cell} instance descriptors $\{\mathcal{F}_{p_i}\}^{N_p}_{i=1}$ into a global cell descriptor $\mathcal{F}_C$ using an EdgeConv layer~\cite{Wang2019TOG} followed by a max pooling operation. 

\PAR{Instance encoder.} 
Each instance $P_i$ is represented by a point cloud where each point contains three spatial and three color (RGB) coordinates, yielding 6D input features (Fig.~\ref{fig:encoders} \textit{bottom}).  
We encode such a point cloud using a PointNet++\cite{Qi2017NIPS} backbone, which gives us a semantic embedding. 
In addition, we explicitly obtain a color embedding of it by encoding its RGB coordinates using our color encoder and a positional embedding of it by encoding its instance center $\Bar{P_i}$, \ie, the mean value of its coordinates, using our positional encoder.
Each of the color encoder and positional encoder takes form of a 3-layer multi-layer perceptron (MLP), whose output dimension is the same as the semantic embedding dimension. 
The semantic, color and positional embeddings are fused by concatenation and fed into a projection layer (another 3-layer MLP) which outputs a final instance embedding $\mathcal{F}_{p_i} \in \mathcal{R}^{D_p}$. 

\subsection{Fine Localization}\label{sec:method_fine}
Given a set of retrieved candidate cells, we now find cues that allow us to refine the position among those cells based on the query description. 
In classical visual localization, such a fine localization step normally relies on establishing 2D pixel correspondences between the query image and retrieved \newtext{\textit{top-k}} database images~\cite{Sarlin2019CVPR, Zhou2020ICRA}. These images have geo-information attached and provide cues to compute an accurate camera position for the query.  
Inspired by this idea, we propose to compute the refined position by establishing hint-to-instance correspondences between a position description and its \newtext{\textit{top-k}} retrieved cells. 

\begin{figure}[t!]
  \centering
  \includegraphics[width=0.98\linewidth]{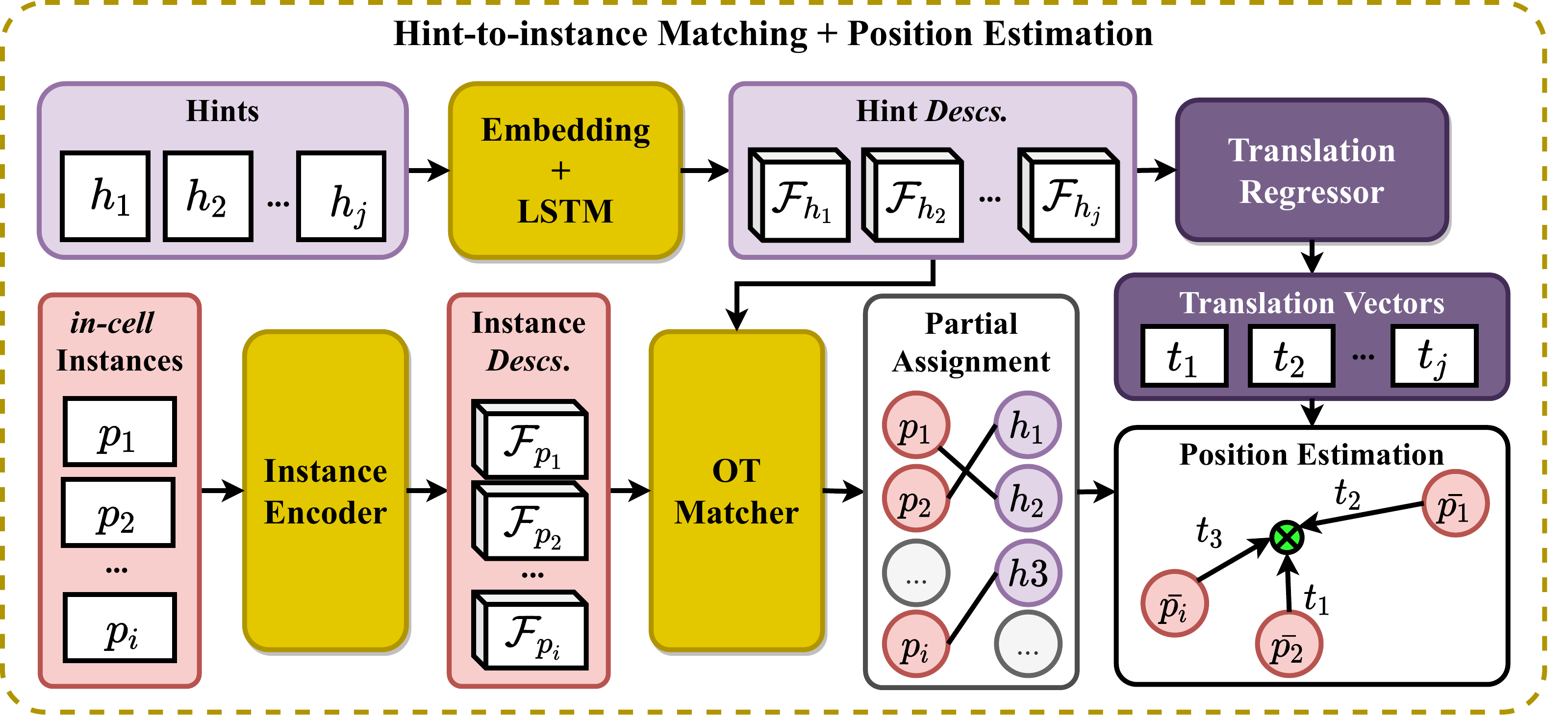}
    \vspace{-7pt}
  \caption{\textbf{Fine localization}. In the fine localization, for each candidate cell, we first establish the correspondences between the query hints and its \textit{in-cell} instances, which allows us to filter out noisy signals that are not useful for position estimation from both domains. We then predict a  vector for each matched instance that translates its instance center to a position estimate via a translation regressor. The final position is the average of the position estimates by all matched instances.}
  \label{fig:fine-loc}
\end{figure}

\PAR{Hint-to-instance matching.}
Given a position description $\mathcal{T}$ and a candidate cell $\mathcal{C}$, we first establish hint-to-instance correspondences with our cross-modal matching module (Fig.~\ref{fig:fine-loc}). 
As explained in Sec.~\ref{sec:dataset}, a position description consists of a series of hints $\{h_j\}^{N_h}_{j=1}$, where each hint describes a single instance about its attributes and relation to the position.
We use our \textit{instance encoder} to extract a descriptor $\mathcal{F}_{p_i}$ for each \textit{in-cell} instance $p_i \in \mathcal{C}$ and our text encoder to extract a descriptor $\mathcal{F}_{h_j}$ per hint. 

We then use a matching module (inspired by SuperGlue~\cite{Sarlin2020CVPR}) to perform a partial matching between the set of instances and the set of sentences. 
The matching module first uses several blocks of self-attention layers and cross-attention layers to propagate contextual information.
The pair-wise similarity scores are computed from the two sets of aggregated descriptors as the cost matrix to an optimal transport (OT) matching layer. 
We adopt the same OT setup as in~\cite{Sarlin2020CVPR}. 
Finally, we obtain the partial assignment from hints to instances by picking matches with confidence scores above the certain threshold ($0.2$ in practice).

\PAR{Translation-based refinement.}  
For each identified match $(p_i, h_j)$ of the instance $p_i$ and the hint $h_j$, we additionally learn a translation vector $t_i$ that transforms the instance center $\Bar{p_i}$ to the target position. 
Each translation prediction $t_i$ then leads to a position estimate that is defined as $\Tilde{y_i}=\Bar{p_i} + t_i$.
To predict such a translation vector, we train an additional 3-layer MLP translation regressor that takes a hint descriptor $\mathcal{F}_{h_j}$ as the input, and outputs a translation vector $t_i$. %
The final position of the query text is computed by taking the average of the set of estimates $\{\Tilde{y_i}\}$ predicted from all hint-to-instance matches. 
We show in our ablation (Sec.~\ref{section:exp_model_ablat}) that our learned translation leads to a more accurate position prediction compared to simple averaging of the matched instance positions. 

\subsection{Losses}
\PAR{Coarse loss.} Given an input batch of cell descriptors $\{\mathcal{F}^i_C\}^{N_b}_{i=1}$ and matching text descriptors $\{\mathcal{F}^i_T\}^{N_b}_{i=1}$ where $N_b$ is the batch size, we train the network for cross-domain retrieval with the pairwise ranking loss~\cite{Kiros2014Arxiv}:
\begin{equation}
\begin{aligned}
    \small
    \mathcal{L}_{coarse} = 
    \sum_{i}^{N_b} \sum_{j \neq i}^{N_b} [\alpha - \langle\mathcal{F}_C^i, \mathcal{F}_T^i\rangle + \langle\mathcal{F}_C^i, \mathcal{F}_T^j\rangle]_{+} \\
    + \sum_{i}^{N_b} \sum_{j \neq i}^{N_b} [\alpha - \langle\mathcal{F}_T^i, \mathcal{F}_C^i\rangle + \langle\mathcal{F}_T^i, \mathcal{F}_C^j\rangle]_{+} \,,
\end{aligned}
\end{equation}
where $[\bullet]_+$ equals $max(0,\bullet)$ and $\alpha$ is the margin hyper parameter.
This loss enforces that each cell descriptor $\mathcal{F}_C^i$ is closer to its matching text descriptor $\mathcal{F}_T^i$ than it is to non-matching text descriptors $\mathcal{F}_T^j$ in the batch by a margin. 
The same is also enforced when matching from a text descriptor $\mathcal{F}_T^i$ to a cell descriptor.

\PAR{Fine loss.} To train our matching module, we adopt the matching loss used in~\cite{Sarlin2020CVPR}, which maximizes the scores at ground truth (GT) match locations in the predicted assignment matrix. We train this model separately from the coarse module. 
To train the translation regressor, we minimize the mean square error between the predicted and the GT translation.
Matching module and the regressor are jointly trained with the sum of the matching loss and regression loss.
We provide implementation details of model training in our supplementary material.

\begin{table}[t!]
  \centering
  \resizebox{\linewidth}{!}{%
  \begin{tabular}{lcccc}
    \toprule
\multirow{2}{*}{Stride} &   \multirow{2}{*}{\# Cells}  & \multicolumn{3}{c}{Localization \newtext{Recall} ($\epsilon<5/10/15m$)}  \\
                &   &   \newtext{$k=1$} & \newtext{$k=5$} & \newtext{$k=10$}  \\
    \midrule
    $\mathcal{S}=10m$                       &  1434     &    \textbf{0.14/0.25/0.31}	&	\textbf{0.36/0.55/0.61}	&	\textbf{0.48/0.68/0.74} \\
    $\mathcal{S}=15m$                       &  629     &    0.10/0.19/0.25	&	0.26/0.47/0.56	&	0.35/0.61/0.70 \\
    $\mathcal{S}=20m$                       &  362     &    0.07/0.15/0.19	&	0.18/0.36/0.45	&	0.25/0.50/0.60 \\
    \bottomrule
  \end{tabular}
  }
    \vspace{-7pt}
    \caption{Ablation on varying sampling stride for cell database construction.}
    \vspace{7pt}
 \label{tab:coarse}
\end{table}

\section{Experimental Evaluation} 
\label{sec:experiments} 

In this section, we discuss the performance of our model (Sec.~\ref{sec:method}) on the proposed \textit{KITTI360Pose} dataset. 
We report results on our validation split for ablation studies (Sec.~\ref{section:exp_model_ablat}) and on our test split for a final evaluation of our best performing models (Sec.~\ref{section:exp_test}).
For details on the dataset and splits, we refer to our supplementary.
\PAR{Evaluation metrics.} We perform the evaluation \wrt \newtext{the top $k$ retrieved candidates ($k \in \{1, 5, 10\}$)} and 
report \newtext{localization recall, \ie, } the \newtext{ratio} of successfully localized queries if its error is below specific error thresholds, \ie, $\epsilon < 5/10/15m$ by default.

\subsection{Model Ablations} \label{section:exp_model_ablat}
\PAR{Database construction.} 
The \textit{coarse localization} module (as explained in Sec.~\ref{sec:method_coarse}) retrieves the \textit{top-k} candidate cells. 
To study its performance \wrt the localization task, we use the center of a cell as a coarse position estimate and measure its accuracy.
As our first ablation, we study the impact of the cell sampling stride $\mathcal{S}$ on localization performance.

As shown in Tab.~\ref{tab:coarse}, the retrieval performance on cells sampled with $\mathcal{S}=10m$ performs better than other stride settings across different \newtext{$k$} values. As the smaller stride implies more overlap between consecutively sampled cells, it helps our model to learn more discriminative descriptors that can be used to distinguish the content of close-by cells. 
While decreasing the stride allows for more accurate localization, it increases the computation demand in terms of memory and runtime in quadratic order. 
We present results for denser sampling using $\mathcal{S}=1/3/5m$ in the supplementary. 
Considering the trade-off between accuracy and computational efficiency, we use $\mathcal{S}=10m$ to train and evaluate our models in the following experiments.

\begin{table}[!t]
  \centering
  \resizebox{\linewidth}{!}{
  \begin{tabular}{llccc}
    \toprule
    \multirow{2}{*}{Train}  &   \multirow{2}{*}{Infer}  & \multicolumn{3}{c}{\newtext{Localization Recall} ($\epsilon<5/10/15m$)}  \\
        &   &   \newtext{$k=1$} & \newtext{$k=5$} & \newtext{$k=10$}  \\
    \midrule
    4    &   6   &   0.12/0.21/0.27	&	0.30/0.48/0.55	&	0.41/0.61/0.67  \\
    6    &   6   &   0.13/\textbf{0.25/0.30}	&	\textbf{0.36/0.54/0.60}	&	\textbf{0.47/0.67/0.73}  \\
    10   &   6   &   \textbf{0.14/0.25/0.30}	&	\textbf{0.36}/0.51/0.57	&	\textbf{0.47}/0.64/0.70  \\
    12   &   6   &   0.11/0.21/0.25	&	0.30/0.45/0.51	&	0.41/0.58/0.64  \\
    \midrule
    6   &   4    &   0.09/0.17/0.22	&	0.26/0.42/0.48	&	0.35/0.55/0.61  \\
    6   &   6    &   0.13/0.25/0.30	&	0.36/0.54/0.60	&	0.47/0.67/0.73  \\
    6   &   10   &   0.20/0.34/0.40	&	0.47/0.68/0.73	&	0.61/0.80/0.84  \\
    6   &   12   &   \textbf{0.23/0.36/0.41}	&	\textbf{0.50/0.69/0.74}	&	\textbf{0.63/0.80/0.84}  \\
    \bottomrule
  \end{tabular}
  }
    \vspace{-7pt}
    \caption{Ablation on the number of hints in a query description.}
    \vspace{7pt}
  \label{tab:hint-ablation}
\end{table}

\begin{table}[!t]
  \centering
  \resizebox{\linewidth}{!}{%
  \begin{tabular}{l c }
    \toprule
    Model &  \newtext{Localization Recall} ($\epsilon<2/5/10m$)   \\
    \midrule
    Center of Cell:    &    0.14/\textbf{0.77/0.99} \\
    Mean of Matched Instance:   &    0.15/0.62/0.97 \\
    Matched Instance + Translation (Text2Pos):    &   \textbf{0.24}/0.76/\textbf{0.99}\\
    \midrule
    Matching Oracle:    &    0.34/0.91/1.00   \\
    Translation Oracle:  &    0.55/0.90/0.99 \\
    Both Oracles:    &   1.00/1.00/1.00 \\
    \bottomrule
  \end{tabular}
  }
    \vspace{-7pt}
   \caption{Ablation on fine localization components. This ablation study on the fine localization module requires narrower localization thresholds compared to Tab.~\ref{tab:coarse} to reveal differences in localization precision.}
      \label{tab:fine}
  \vspace{7pt}
\end{table}

\PAR{Ablation on number of localization cues.}
To evaluate how the number of specified object cues influences localization \newtext{recall}, we vary the number of hints $N_h$ from four to 12 either during training or inference, as shown in Tab.~\ref{tab:hint-ablation}. The results show that alternating the number of hints below or above $6$ during training decreases the \newtext{recall} in most metrics. Furthermore, we show that our model trained on $N_h=6$ hints is robust to varying $N_h$ during inference and that performance even rises by up to $16$ points for $N_h=12$, which is to be expected as additional hints help to resolve some of the ambiguity in cross-modal localization.

\PAR{Fine localization module.} 
In this ablation, we provide insights into design decisions for the \textit{fine localization} module~(Sec.~\ref{sec:method_fine}). 
To make the evaluation independent of the retrieval module performance, we use a retrieval oracle in the following experiments to provide the model with the database cell that is closest to the GT position.

\textbf{Firstly}, we study the performance of the hints-to-instance matching module, by measuring precision and recall of the predicted matches given the GT matches.
Our matching model can achieve 78\% and 76\% for both recall and precision, respectively. 

\textbf{Next}, we show the benefit of our translation regressor by comparing the following three different variants for position estimation: (i) we take the cell-center as the estimated position (no refinements); (ii) we compute the mean of the instances matched by query hints, and (iii) we use the mean of the position estimates computed by using the predicted translation vectors.
Due to the use of a retrieval oracle, we report their localization \newtext{recall} with stricter error thresholds $\epsilon < 2/5/10m$.
As shown in Tab.~\ref{tab:fine} (\textit{top}), all three variants can localize queries within $10m$ errors with almost $100\%$ \newtext{recall}. 
The difference between model performance becomes more evident when using the smaller error threshold $\epsilon=2m$, where we see the simple cell-center baseline performs poorly to localize only 14\% of queries. This only improves marginally by 1\% if we take the mean center of matched instances. Additionally, we even see the opposite results for a moderate error of $\epsilon=5m$ where the cell-center baseline outperforms the instance-mean baseline by 15\%. 
The results thus suggest that naively assuming either the position to be at the center of the cell or between the described instances is not sufficient.
Instead of making such manual assumptions, we learn where to expect the position given an instance of a specific object class in a data-driven manner.
We show that our learned translation leads to 24\% localization \newtext{recall} on the smallest threshold, outperforming the other two variants by up to 10\% and confirming its potential efficacy for fine-grained localization.

\textbf{Finally}, we replace parts of the \textit{fine model} by oracles to understand the limitations of individual components of our method. 
As shown in Tab.~\ref{tab:fine} (\textit{bottom}), using GT associations (\textit{matching oracle}) instead of performing hint-to-instance matching using our model, we localize ($10\%/15\%$) more queries within $2/5m$ errors. 
Purely replacing predicted translation with GT translations (\textit{translation oracle}) leads to $31\%$ more queries localized within $2m$, suggesting room for improvements by predicting more accurate translation vectors.
Both oracles combined yield perfect localization as expected.

\PAR{Pipeline-level oracle ablation.} 
While the previous experiments used a coarse oracle to evaluate fine localization in isolation, we now replace first our coarse and then our fine module by oracles, in order to understand their limitations of the full system. The results are shown in Tab.~\ref{tab:oracle}.
With \textit{coarse oracle}, we replace the retrieval component with a retrieval oracle (as in the previous fine module ablation). 
As can be seen, identifying cells reliably localizes all queries within $15m$ distance to the query position, and with $98\%$ accuracy within $10m$ threshold. This clearly shows significant potential in improving the retrieval module, which can be achieved potentially with additional information (such as nearby landmarks or street names) to reduce the inherent cell ambiguity. This remains our future work. 
With \textit{fine oracle}, we then measure the localization performance of our learned retrieval model + GT matching association + GT translation prediction.
We show that perfect fine localization improves the baseline by 11\%/29\%/31\%  at the smallest thresholds for \newtext{$k=1/5/10$} candidate cells, which indicates further potential in improving the refinement module.
Finally, using both oracles, we can localize all queries within $5m$ distance from the query position.
\newtext{For the lower part of Tab.~\ref{tab:oracle}, we also replace our coarse and fine modules with random choice to highlight their importance. Compared to our baseline, a random coarse retrieval leads to near-zero localization recall,  while a random matching in the fine module leads to localizing up to $34$ \% less queries.}

\PAR{Comparison to visual localization}
While Text2Pos is not intended as a competitor against visual localization, we do provide a comparison in localization \newtext{recall} through a two-step experiment (Tab.~\ref{tab:vis-loc}).  Visual localization has to be performed on rendered images in order to be applicable on our dataset, so we first show a comparison between real and rendered images from identical locations, confirming that rendered images do not significantly reduce retrieval performance. 
Then, we compare visual retrieval (NetVLAD~\cite{Arandjelovic2016CVPR}\footnote{We use the pretrained NetVLAD obtained from \href{https://github.com/Nanne/pytorch-NetVlad}{here}}) to our cross-modal retrieval by rendering images at each cell and query pose of our validation set, resulting in $5736$ and $3187$ images on the database and query side, respectively.

The results indicate that visual retrieval shows superior accuracy for smaller \textit{top-k} values with up to $13$ percentage points at $k=1$ and $\epsilon = 15m$, but also lags behind cross-modal retrieval with up to $6$ percentage points in $k=10$ and $\epsilon = 10m$. We note that for simplicity, this comparison considers retrieval only without refinement and that more advanced visual localization pipelines are available. Nevertheless, we take this as a first indication that equally strong advancements in cross-modal localization might approach the accuracy of state-of-the-art visual pipelines.

\begin{table}[!t]
  \centering
    \resizebox{\linewidth}{!}{%
  \begin{tabular}{lccc}
    \toprule
    \multirow{2}{*}{Model}  & \multicolumn{3}{c}{\newtext{Localization Recall} ($\epsilon<5/10/15m$)}  \\
         &   \newtext{$k=1$} & \newtext{$k=5$} & \newtext{$k=10$}  \\
    \midrule
    Text2Pos                    &   0.14/0.25/0.31	&	0.36/0.55/0.61	&	0.48/0.68/0.74 \\
    \midrule
    Coarse Oracle               &   0.68/0.98/1.00	&  -	&   - \\
    Fine Oracle                 &   0.35/0.37/0.38	&   0.65/0.67/0.68	&   0.77/0.78/0.80 \\
    Both Oracles                &   1.00/1.00/1.00	&   1.00/1.00/1.00  &	1.00/1.00/1.00 \\
    \midrule
    Coarse random        &   0.00/0.01/0.01 &	0.01/0.02/0.03 &	0.02/0.04/0.06 \\   
    Fine random          &   0.03/0.10/0.20    &	0.09/0.30/0.49 & 0.14/0.40/0.62 \\   
    \bottomrule
  \end{tabular}
  }
    \vspace{-7pt}
   \caption{Ablation on localization oracle. \newtext{Note that the line 2 results differ from Tab.~\ref{tab:fine} line 3 due to different experimental setups.}}
    \label{tab:oracle}
    \vspace{7pt}
\end{table}

\begin{table}[t!]
  \centering
    \resizebox{\linewidth}{!}{%
  \begin{tabular}{lccc}
    \toprule
    \multirow{2}{*}{Model}  & \multicolumn{3}{c}{\newtext{Localization Recall} ($\epsilon<5/10/15m$)}  \\
        & \newtext{$k=1$} & \newtext{$k=5$} & \newtext{$k=10$}  \\
    \midrule
    NetVLAD (real)        &   0.11/0.51/0.79 &	0.11/0.59/0.95 &	0.11/0.60/0.96 \\
    NetVLAD (rendered)        &   0.10/0.49/0.74 &	0.11/0.59/0.93 &	0.11/0.59/0.95 \\
    \midrule
    Text2Pos (w/o fine loc.) &   0.10/0.23/0.30 &  0.27/\textbf{0.52}/0.60 &  \textbf{0.37/0.65/0.72} \\
    NetVLAD~\cite{Arandjelovic2016CVPR}  &   \textbf{0.18/0.33/0.43} &  \textbf{0.29}/0.50/\textbf{0.61} &  0.34/0.59/0.69 \\
    \bottomrule
  \end{tabular}
  }
    \vspace{-7pt}
   \caption{Comparison with visual localization. A pre-trained NetVLAD model for visual retrieval is compared between real and rendered images (top rows) and against our novel cross-modal retrieval.}
   \label{tab:vis-loc}
   \vspace{7pt}
\end{table}

\subsection{Evaluation on the Test Set}\label{section:exp_test}

\begin{table}[t!]
  \centering
    \resizebox{\linewidth}{!}{%
  \begin{tabular}{lccc}
    \toprule
    \multirow{2}{*}{Model}  & \multicolumn{3}{c}{\newtext{Localization Recall} ($\epsilon<5/10/15m$)}  \\
        & \newtext{$k=1$} & \newtext{$k=5$} & \newtext{$k=10$}  \\
    \midrule
    Text2Pos (full)    &  \textbf{0.13/0.21/0.25}  &   \textbf{0.33/0.48/0.52}   &   \textbf{0.43/0.61/0.65}   \\
    Text2Pos (w/o trans. offs.)    &  0.10/0.20/\textbf{0.25}   &   0.26/0.46/\textbf{0.52}   &   0.35/0.58/\textbf{0.65}  \\        
    Text2Pos (w/o fine loc.)    &   0.10/0.20/\textbf{0.25}  &   0.27/0.46/\textbf{0.52}  &   0.35/0.59/\textbf{0.65} \\
    \bottomrule
  \end{tabular}
  }
    \vspace{-7pt}
   \caption{Model evaluation on the test set.} 
      \label{tab:components}
    \vspace{7pt}
\end{table}

As can be seen in Tab.~\ref{tab:components}, for top-1 retrieval, we can successfully localize up to $25\%$ of queries up to $15m$ distance threshold and $20\%$ of queries up to a distance of $10m$. 
When considering top-5 queries, we can already successfully localize $52\%$ of all queries and go up to $65\%$ successfully localized queries when considering top-10 cell candidates. 
As can be seen, based on purely textual descriptions, we can already localize the target position quite accurately (\ie, $10m$ radius). However, to get to a good localization \newtext{recall}, we need to consider top-10 cell candidates. 
This suggests that our model is rather uncertain which cell contains the query position. This is not surprising, as many of the query descriptions could refer to several locations in the city. Additional hints such as a nearby address, nearby street names, or landmarks should help in the future to further narrow down the search space and resolve this ambiguity and provide a good direction for future research.

Finally, as can be seen in Tab.~\ref{tab:components}, the full localization variant using translation prediction outperforms the variant using naive matched-instances position averaging by localizing 3\% more queries within $5m$ errors considering top-1 retrieved cell. This improvement increases to 8\% when considering top-10 candidates.

\section{Broader Impact and Limitations}

Our work on text-based localization opens a new front of research on natural-language-based action coordination with future mobile systems, to which we may need to specify our current or a target location. Usage scenarios include autonomous goods delivery (\eg, food or packages), ordering autonomous vehicle pick-up, or sending vehicles to remote locations in case of emergency. These use-cases will play an important role in the automation of tasks that can be considered unsafe (due to a high number of traffic accidents) and currently require a human operator. Automation of these tasks can also improve the weight and utilization of delivery vehicles and, consequentially, the carbon footprint. 

\newtext{However, to devise a first feasible solution, we rely on the following assumptions: (i) we assume our maps contain labeled instances of objects used as anchor points for localization; (ii) we rely on simplistic, template-based localization instructions for training and evaluation. Generalization to arbitrary, unlabeled point clouds and more realistic, human-generated textual queries remains our future work.} 

\section{Conclusion}

We presented Text2Pos and KITTI360Pose, the first method and dataset for text-based position localization within a 3D environment. 
As such language-based communication is natural to human beings, we foresee it will be an integral part of future mobile agents that will require location instructions, such as goods delivery or vehicle pick-up positions. We demonstrated that our coarse-to-fine approach can localize $65\%$ of textual queries within $15m$ distance to query locations when considering top-10 retrieved locations. 
We believe we can further improve localization precision by using street names and visual landmarks as cues for coarse localization, which we leave for future work. This work is the first step in the direction of language-based localization, showing its great potential, and hopefully inspiring researchers to further make this technology a reality.

\footnotesize{\PAR{Acknowledgments.} This project was partially funded by the
Sofja Kovalevskaja Award of the Humboldt Foundation. The authors of this work take full
responsibility for its content.

\clearpage
\appendix

\noindent In this supplementary material, we provide additional information to further understand our proposed coarse-to-fine language-based localization model -- Text2Pos. 
In Sec.~\ref{supp:dataset}, we provide more details about our \textit{KITTI360Pose} dataset such as the query position sampling mechanism and the clustering and describing of scene object instances.
Next, we describe implementation details about the cell database construction, data augmentation, model training and PointNet++ pre-training in Sec.~\ref{supp:implement_details}.
We further present thorough ablation experiments to study the impact of data variations on the localization performance in Sec.~\ref{supp:ablation}.
There we also prove our concept of utilizing street names (when available) to better overcome the ambiguity in descriptions caused by the fact that many locations have semantically similar or equal surroundings.
Finally, Sec.~\ref{supp:qualitative} shows qualitative results of top-k candidate cells retrieved during the coarse localization stage.
We will release our code upon the paper's acceptance.

\section{KITTI360Pose Dataset Details}\label{supp:dataset}
\PAR{Query position sampling and description.}
To obtain query positions, we first sample equidistant locations along the original KITTI360~\cite{Xie2016CVPR} capturing paths. For each of these trajectory locations, we then sample a number of random nearby positions that are up to \nicefrac{1}{2} cell-sizes away from it.
We sample 4 positions for the baseline and present detailed studies on the impact of training our baseline on a larger dataset that samples more query positions (per trajectory location) in Tab.~\ref{tab:larger-dataset}. 

As detailed in the main document, a query position is described based on the set of its surrounding object instances, and thus we discard a sampled query position if there are insufficient instances, \ie $N_h < 6$ in practice, in its vicinity.
Given a query position with enough nearby instances, we can select the subset of the instances (6 instances in practice) to describe the query using three different strategies: (i) simply selecting the closest-by instances, (ii) choosing the instances that cover as many different directions (pointing towards the query position) as possible and (iii) picking instances with as many different classes as possible.
If two or more of these strategies yield the same set of instances, the duplicates are omitted.

\PAR{Object clustering.}
In order to also incorporate stuff class objects into our descriptions, it is necessary to cluster them into separate instances, \eg, clustering one large \textit{sidewalk} object covering the entire scene into separate instances left and right of a query position.
Since we did not find a way to achieve such a clustering globally for a whole scene, we decide to cluster a stuff-class object locally for a given cell, where we first ignore those points of the complete stuff object that are not contained in that cell.
For its points inside the cell, we then cluster them into multiple separate instances using the DBSCAN~\cite{Ester96KDD} and retain the clustered instances that contain enough points (250 in practice).

\PAR{Dataset Statistics.}
As described above, we can obtain datasets of different scales by varying the number of sampled positions at each trajectory location. 
In Tab.~\ref{tab:data-stats}, we show statistics about the number of query positions, descriptions and covered scene area of the two \textit{Kitti360Pose} dataset versions generated by sampling 4 and 8 positions per trajectory location.

\begin{table*}[t]
  \centering
  \begin{tabular}{lccccc}
    \toprule
    Split   &   \# Scenes   &   Area [$km^2]$   &   \# Positions   &   \# Descriptions & \# Unique Desc.\\
    \midrule
    \multicolumn{5}{l}{4 sampled position per trajectory location}    \\
    \midrule
    Training   &   5   &   11.59   &   9961   &   28807     &   601   \\
    Validation  &   1   &   2.25    &   1116   &   3187     &   416 \\
    Testing     &   3   &   2.14    &   3932   &   11505    &   518   \\
    All         &   9   &   15.98   &   15009   &   43499   &   629   \\
    \midrule
    \multicolumn{5}{l}{8 sampled position per trajectory location}    \\
    \midrule
    Training   &   5   &   11.59   &   19688   &   57482    &    612\\
    Validation  &   1   &   2.25    &   2224   &   6405     &    435\\
    Testing     &   3   &   2.14    &   7747   &   22878    &    542\\
    All         &   9   &   15.98   &   29659   &   86765   &    641\\
    \bottomrule
  \end{tabular}
    \vspace{-7pt}
    \caption{\textit{Kitti360Pose} dataset statistics}
    \vspace{7pt}
  \label{tab:data-stats}  
\end{table*}

\PAR{Challenges.}%
Notice, the mapping between a query description and a target position is often not unique, since there exist many positions with semantically similar or even equal surroundings.
Such ambiguous nature makes the task of text-based outdoor localization very challenging.
Compared to the indoor environment which is rich in semantics of various objects such as table, sofa, chair and cup ~\cite{Dai2017CVPR}, outdoor scenes have fewer static objects, yet more complicated semantic cues that are mainly based on \textit{stuff} classes, such as tree, vegetation, fence and road.
Those \textit{stuff}-based semantic features are highly repetitive for real-world large-scale outdoor scenes, which is also verified by our statistics in Tab.~\ref{tab:data-stats}, which show that only a small fraction of the generated descriptions is actually unique.
Therefore, our method is designed to endure such ambiguity by retrieving several candidate cells and then performing more refined pose estimations for each of them.
In future work, we expect to reduce the ambiguity by incorporating the use of unique landmarks like street names or named buildings, which can be easily integrated into our general coarse-to-fine localization pipeline. 
We present our experiment in Tab.~\ref{tab:street-names} as a proof of concept, where we manage to utilize simulated street names to improve our localization performance up to 20 percent points.

\section{Implementation Details}\label{supp:implement_details}

\PAR{Cell database.}
Our method relies on a database of cells to first retrieve top-k candidate cells which potentially contain our target position and then perform more accurate fine localization within those cells. 
To construct a database of cells that can fully cover the scene area, we use a sliding window with size $\mathcal{W}$ and sliding stride $\mathcal{S}$ to sample the cells along both the horizontal and vertical directions. 
We empirically fix the cell size to be $\mathcal{W}=30m$ which usually covers enough instances for our experiments and we use a stride of $\mathcal{S} = \nicefrac{1}{3} \times \mathcal{W}$.
After the raw sampling, we further reject cells that have not enough objects inside to describe a position, \eg a cell with mostly empty space.
This leads to a database of $11259/1434/4308$ cells for the training/validation/testing scene split and in total $17,001$ cells for the whole dataset.

\PAR{\textit{In-cell} instances.} For fine localization, we cut-off or pad \textit{in-cell} instances to keep the same number of instances per cell to allow mini-batching in the matching module. We consider each cell to contain 16 \textit{in-cell} instances, \ie, $N_p = 16$ for both training and inference, which has led to the most promising performance across different data settings in our experiments.
In the case of too few instances, we pad dummy instances of 10 points which have black color and random point coordinates close to zero so that they are easily recognized by the model.
In addition, we normalize the coordinates of \textit{in-cell} instances \wrt the size of their belonging cell such that each coordinate value is $\in [0,1]$, which aids the regressor's training stability.

\begin{table*} [t]
  \centering
  \begin{tabular}{lcccc}
    \toprule
\multirow{2}{*}{Stride} &   \# Cells  & \multicolumn{3}{c}{\newtext{Localization Recall ($\epsilon<5/10/15m$)}}  \\

        &  (val split) &  \newtext{$k=1$} & \newtext{$k=5$} & \newtext{$k=10$}  \\
    \midrule
    $\mathcal{S}=3m$                        &  15899    &    \textbf{0.22/0.35/0.41}	    &	\textbf{0.41}/0.53/0.58	&	\textbf{0.52}/0.63/0.68 \\
    $\mathcal{S}=5m$                        &  5724     &    0.18/0.29/0.35     &   0.39/0.53/0.57  &   0.51/0.64/0.68 \\
    $\mathcal{S}=10m$ (baseline)            &  1434     &    0.14/0.25/0.31	    &	0.36/\textbf{0.55/0.61}	&	0.48/\textbf{0.68/0.74} \\
    $\mathcal{S}=15m$                       &  629      &    0.09/0.19/0.25	    &	0.25/0.46/0.54	&	0.35/0.61/0.70 \\
    $\mathcal{S}=20m$                       &  362      &    0.07/0.14/0.19     &   0.19/0.37/0.45  &	0.25/0.49/0.59 \\
    \bottomrule
  \end{tabular}
    \vspace{-7pt}
    \caption{Ablation on varying sampling stride for cell database construction.}
    \vspace{7pt}
\label{tab:cell-stride}  
\end{table*}

\begin{table}[t!]
  \centering
\resizebox{\linewidth}{!} {      
  \begin{tabular}{lccc}
    \toprule
    Positions  & \multicolumn{3}{c}{\newtext{Localization Recall} ($\epsilon<5/10/15m$)}  \\
    (per traj. loc)   & \newtext{$k=1$} & \newtext{$k=5$} & \newtext{$k=10$}   \\
    \midrule
    4 (Baseline) &  0.14/0.25/0.31	&	0.37/0.54/0.60	&	0.48/0.68/0.73 \\
    8    &  0.16/0.28/0.33	&	0.39/0.57/0.63	&	0.52/0.70/0.75 \\    
    12  &  0.16/0.29/0.35	&	0.41/0.60/0.66	&	0.52/0.72/0.77 \\    
    16   &  0.16/0.28/0.34	&	0.39/0.56/0.62	&	0.51/0.70/0.75 \\    

    \bottomrule
  \end{tabular}
  }
    \vspace{-7pt}
    \caption{Training on larger dataset. We generate a larger dataset for training by varying the number of positions sampled at every trajectory location.}
    \vspace{7pt}
  \label{tab:larger-dataset}  
\end{table}

\PAR{Query description grounding.}
To learn the task of text-to-cell retrieval, we need to ground a query position description onto a ground-truth (GT) cell.
We define a GT cell to be the cell in the database that contains the described position and whose center is closest to the described position.
To learn the task of hint-to-instance matching, we need to further identify the GT correspondences between the query hints and the \textit{in-cell} instances within a candidate cell.
For \textit{instance}-class instances, the hint-to-instance correspondences are established using GT instance IDs from the original KITTI360 dataset.
Notice, \textit{stuff}-class hint instances were clustered using a synthetic cell centered on the query position, which means the same instance might appear in a slightly different position within a retrieved cell since the query position can be anywhere inside that retrieval cell.
Therefore, to match between a \textit{stuff}-class hint instance and a point-cloud \textit{in-cell} instance which are clustered \wrt a synthetic cell and a retrieval cell, we compare their semantic class IDs and the two direction vectors pointing from the query position to those two instances.
We consider them as a match, if they have the same semantic class and their direction vectors are close enough.

\PAR{Data augmentation.} 
During training, we also use data augmentations for the point cloud instances and the pose descriptions to aid our model performance. In the \textit{instance augmentation}, we randomly rotate the instance across the z-axis during training and normalize-scale all its points to a $[0,1]$ interval during the training and inference. 
For the \textit{description augmentation}, we (i) randomly shuffle the order of the hints that make up a query description and (ii) randomly flip cells horizontally and/or vertically by flipping the location of the pose and instance centers in each cell and changing the corresponding words in the description. 
The description augmentations are not used when training the refinement module. 

\PAR{Text2Pos model training.} We train our model using five training scenes and use one scene for model validation.
For coarse localization, we train the retrieval model using an Adam optimizer~\cite{Kingma2015ICLR} with batch size 64 and learning rate 0.001 for at most 64 epochs.
We set the margin parameter $\alpha$ in the pairwise ranking loss (Eq. (1)) as $\alpha = 0.35$.
For fine localization, we train the matching module and regression jointly using an Adam optimizer with batch size 32 and learning rate 0.0003 for 16 epochs.

\PAR{PointNet++ pretraining.}
We pre-train our PointNet++~\cite{Qi2017NIPS} backbone for the point cloud classification on KITTI360 and use it to initialize the two instance encoders used in the coarse and fine localization stages. We aggregate the objects from all our database-side cells into a training set of $159,828$ objects from 22 classes and again use the Adam optimizer~\cite{Kingma2015ICLR} with a learning rate of $0.003$ and a batch size of 32.

\section{More Ablation Studies}\label{supp:ablation}
In the main document, we have presented several ablation studies that focus on analysing the localization performance of each component of our proposed Text2Pos model, \ie, the coarse retrieval component and the fine matching component. 
In this section, we first complete the results of the cell stride ablation (that has been presented in Tab.~\ref{tab:coarse} of the main paper).
And we further provide an additional ablation to study the impact of training on a larger version of our dataset on the localization performance, to thoroughly understand our proposed method.
Finally, we confirm our hypothesis that our localization performance can be further improved when provided with additional landmarks information such as street names.

\begin{figure*}[h!]
  \centering
  \includegraphics[width=0.95\textwidth]{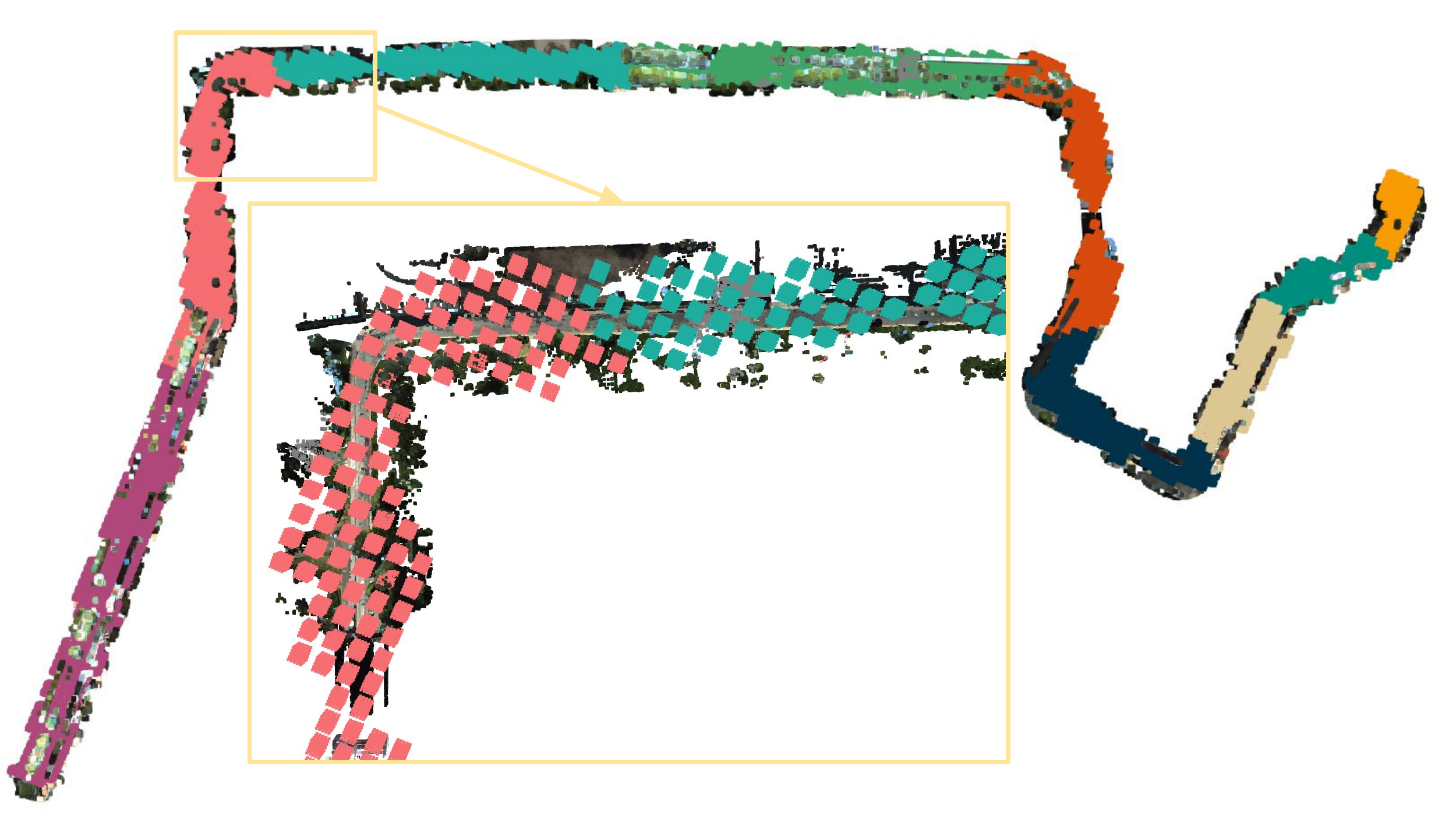}
  \caption{Simulated streets.   \textit{Overview:} We split our validation scenes into nine hypothetical streets marked with different colors. \textit{Zoom-in:} Each colored square placed on top of a street represents a retrieval cell belonging to that street area.}
  \label{fig:streets}
\end{figure*}

\PAR{Full results on sliding stride ablation.}
We variate the stride size for our database-side cells between values decreasing from $\mathcal{S}=20m$ down to $3m$, while keeping a fixed cell size of $30m$. 
As shown in Tab.~\ref{tab:cell-stride} where we mark the best \newtext{recall} with \textbf{bold}, our baseline setting $\mathcal{S}=10m$ leads to the best performance at $10/15m$ error thresholds when using top-5/10 candidates. 
For $5m$ errors or top-1 candidates, the \newtext{recall} steadily increases for smaller strides, \eg, from $7\%$ up to $22\%$ \newtext{recall} for a top-1 candidate prediction within a $5m$ error threshold.
However, we consistently observe a larger performance drop (if any) when going from $\mathcal{S}=15m$ to $10m$ than going from $\mathcal{S}=10m$ to $5m$.
While, $\mathcal{S}=3m$ is the most promising one for the finest error threshold, it leads to less efficient computation in both memory and run time due to the large amount of generated cells (11 times more than in the $10m$ setting).
For our current implementation, the clustered stuff instances are pre-computed and stored separately for each cell, meaning it takes approximately 9 GB of memory to store 15k cells (0.6MB per cell) just for the validation split.
Another option to avoid big memory consumption is to perform instance clustering on-the-fly during training and inference, however, loading plus clustering of all cells in the validation split already takes around $50$ minutes ($0.2s$ per cell).
This also prevents us from pushing the stride to more extreme settings, \eg,  $\mathcal{S}=1m$ where one needs to handle approximately $135k$ cells for the validation split.
To maintain the feasibility of performing intensive evaluations, we choose our baseline setting as $\mathcal{S}=10m$. This gives us the best trade-off between computational efficiency and localization \newtext{recall}, as we consider further technical engineering to improve the computational overhead coming from instance clustering out of the scope of this research, leaving it to future work.

\begin{table*}[t!]
  \centering
\resizebox{0.7\linewidth}{!} {      
  \begin{tabular}{lccc}
    \toprule
    Variant & \multicolumn{3}{c}{\newtext{Localization Recall ($\epsilon<5/10/15m$)}}  \\
            & \newtext{$k=1$} & \newtext{$k=5$} & \newtext{$k=10$}  \\
    \midrule
    Text2Pos (coarse)      & 0.10/0.23/0.30    & 0.27/0.52/0.60    &	0.37/0.65/0.72 \\
    Text2Pos (coarse + fine)    & 0.14/0.25/0.31	& 0.37/0.54/0.60    &	0.48/0.68/0.73 \\
    \midrule
    Text2Pos + street names (coarse)  & 0.15/0.34/0.45	& 0.40/0.69/0.79	& 0.52/0.83/0.89 \\
    Text2Pos + street names (coarse + fine)& \textbf{0.19/0.36/0.46}	& \textbf{0.47/0.72/0.80}	& \textbf{0.61/0.85/0.90} \\
    \bottomrule
  \end{tabular}
  }
    \vspace{-7pt}
    \caption{Using street names as additional localization cues.}
    \vspace{7pt}
  \label{tab:street-names}  
\end{table*}

\PAR{Using a larger dataset.}
As described before, it is possible to increase our dataset scale by sampling more poses around each trajectory location. 
In our final experiment, we vary this scale by sampling 8, 12 and 16 poses per location and train our coarse and fine models on these larger datasets. 
To maintain the comparability, we evaluate models trained under varying settings on the same validation set that is obtained using the baseline configuration.
As shown in Tab.~\ref{tab:larger-dataset}, localization \newtext{recall} improves incrementally by training on a larger dataset, acquired by sampling poses from 4 to 12 positions per trajectory location.
However, the performance starts to decrease on the 16-positions sampling configuration, which indicates a limit of performance improvement by sampling denser query positions.
Our intuition is that the increase in sampling density leads to increases in the ambiguity of the mapping between a query description and a position, which starts to harm the learning of the global mapping on the cell level and the local mapping on the instance level.
We note that we chose the sub-optimal setup with 4 poses per location as our experimental baseline due to time constraints, but will publish our results based using a more optimal sampling in our camera ready version.

\PAR{Using street names as additional cues.}
As mentioned in the main paper as well as Sec.~\ref{supp:dataset}, our method can potentially use additional information such as nearby landmarks or street names to reduce the inherent cell ambiguity caused by places with semantically similar surroundings.
To confirm this hypothesis we perform an additional experiment in which we use simulated street names as our additional cues to improve our localization \newtext{recall}.
To simulate street names, we split our validation scene into nine separate streets as shown in Fig.~\ref{fig:streets} (each street in a unique color).
All database-side cells and query-side poses are then assigned to their corresponding streets. 
During inference, we still perform coarse retrieval as usual, but additionally reject all cells from incorrect streets before taking the \textit{top-k} retrievals. 
Hence, this experiment serves as a "semi-oracle" for the coarse retrieval.
As Tab.~\ref{tab:street-names} shows, using the simulated street names (as prior ground truth) boosts the performance of our baselines up to $20$ percentage points.
With this improvement, our full pipeline is able to localize $19\%$ of the queries with top-1 retrieval within a strict error radius of $\epsilon = 5m$.
If we consider all top-10 retrievals, it then manages to localize $61/85/90\%$ of the queries at different error thresholds of $5/10/15m$. 
Therefore, our experimental results indicate that additional text-based cues like street names are valuable information to complement the challenge of description ambiguity.
We believe our proof of concept shows this direction is worthy of further research investigation to properly incorporate existing landmark information such as street names, zip codes or named buildings, into text-based localization pipelines for performance improvement.

\begin{figure*}[t!]
  \centering
  \includegraphics[width=1\textwidth]{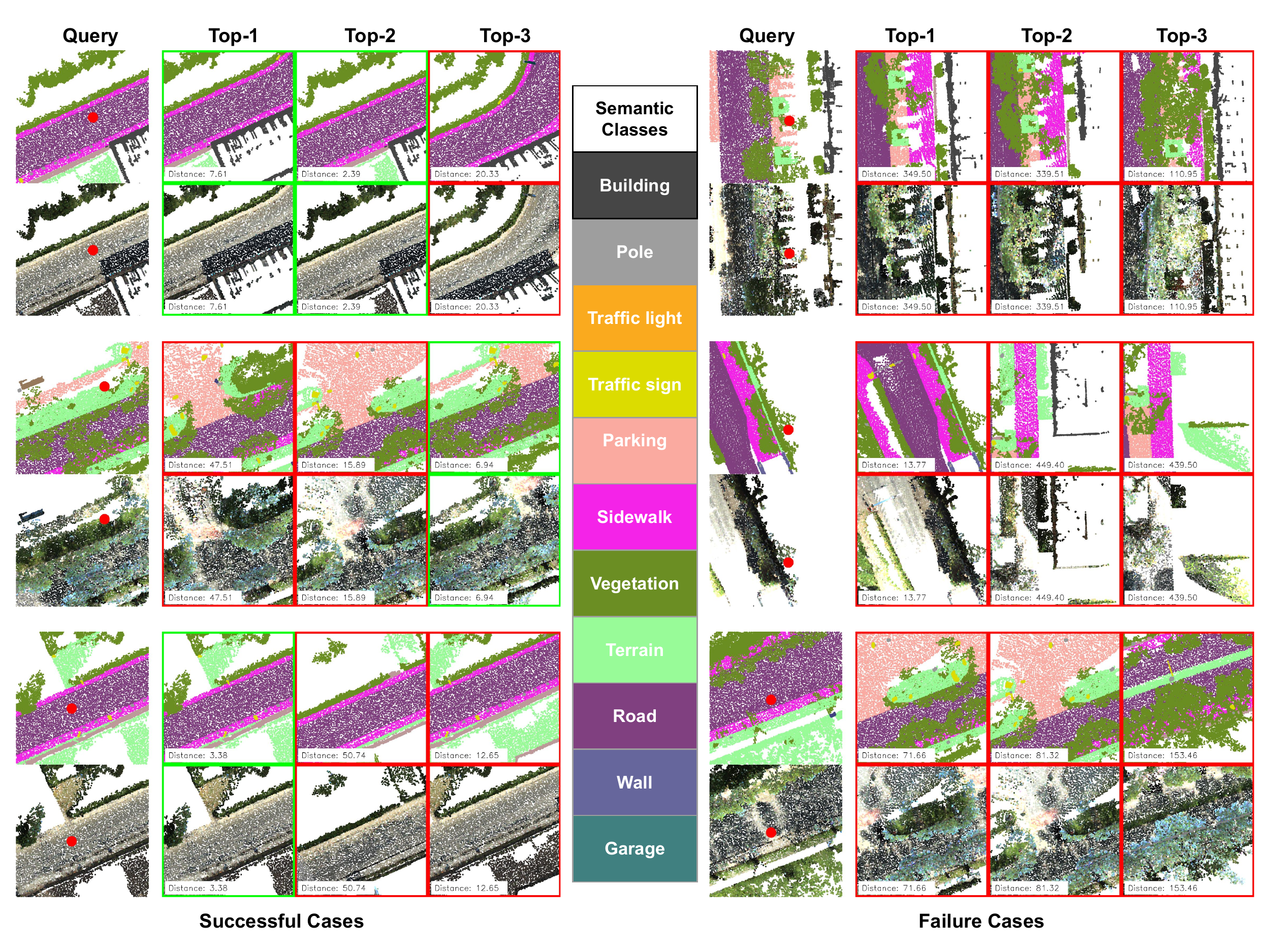}
  \caption{\textbf{Examples of the top-3 retrieved cells.} In the \textit{left} part of the figure, we show 3 successful examples where the correct cell is within the top-3 candidates. In the \textit{right} part, we show 3 failure cases where none of the top-3 candidates is the correct one.
  Each example consists of the dataset cell that is closest to the query pose (the 1st column) and the top-3 retrieval cells, where the $in-cell$ instances are colored by their semantic classes (the top row) and RGB values (the bottom row).
  We plot the mapping from semantic colors to class labels in the center of the figure.
  We further mark the correct cells with \textcolor{green}{green} borders and wrong cells with \textcolor{red}{red} borders, where we consider a  cell to be correct if its center is at most $10m$ away from the query position (the red dot in the query cell).
  }
  \label{fig:retrieval}
\end{figure*}

\section{Qualitative Text-to-Cell Retrieval Analysis}\label{supp:qualitative}
Finally, we show examples of the top-3 candidate cells retrieved by our text-to-cell retrieval model in Fig.~\ref{fig:retrieval}.
As we can see in both the successful (\textit{left}) and failure cases (\textit{right}), top-3 retrieved cells often exhibit high similarity in semantics to the query cell and to each other, yet can be more than $100m$ apart, which again highlights the inherent ambiguity of large-scale outdoor text-based localization. 
As a consequence, as shown in the 2nd successful example (\textit{left}), the correct cell is retrieved but ranked lower (as the 3rd) than the other two candidates which are much further away.
Furthermore, the 2nd failure example (\textit{right})) shows that the top-1 candidate is indeed a close-by candidate yet will not be considered as a correct one according to our manual $10m$ threshold, meaning the pose error is too coarse.
This also suggests the need of another refinement step to improve the localization \newtext{recall}, which will turn such a case into a successful localization.


{\small
\bibliographystyle{ieee_fullname}
\bibliography{references}
}

\end{document}